\documentclass[10pt,twocolumn,letterpaper]{article}
\pdfoutput=1
\usepackage{iccv}
\usepackage{times}
\usepackage{epsfig}
\usepackage{graphicx}
\usepackage{amsmath}
\usepackage{amssymb}

\usepackage[lined,boxruled,linesnumbered]{algorithm2e}
\usepackage{subfigure}
\usepackage{authblk}

\usepackage[breaklinks=true,bookmarks=false]{hyperref}

\iccvfinalcopy 

\def\iccvPaperID{****} 

\ificcvfinal\fi
\begin{document}

\title{Robust Performance-driven 3D Face Tracking in Long Range Depth Scenes}

\setcounter{Maxaffil}{3}

\author[1]{\rm Hai X. Pham}
\author[2]{\rm Chongyu Chen}
\author[3]{\rm Luc N. Dao}
\author[1]{\rm Vladimir Pavlovic}
\author[3]{\rm Jianfei Cai}
\author[3]{\rm Tat-jen Cham}

\affil[1]{Department of Computer Science \\ Rutgers University \\ USA}
\affil[ ]{\textit {\{hxp1,vladimir\}@cs.rutgers.edu}}
\affil[2]{Department of Computing \\ Hong Kong Polytechnic University \\ Hong Kong}
\affil[ ]{\textit {cscychen@comp.polyu.edu.hk}}
\affil[3]{School of Computer Engineering \\ Nanyang Technological University \\ Singapore}
\affil[ ]{\textit {\{nldao,asjfcai,astfcham\}@ntu.edu.sg}}


\maketitle

\begin{abstract}
We introduce a novel robust hybrid 3D face tracking framework from RGBD video streams, which is capable of tracking head pose and facial actions without pre-calibration or intervention from a user. In particular, we emphasize on improving the tracking performance in instances where the tracked subject is at a large distance from the cameras, and the quality of point cloud deteriorates severely. This is accomplished by the combination of a flexible 3D shape regressor and the joint 2D+3D optimization on shape parameters. Our approach fits facial blendshapes to the point cloud of the human head, while being driven by an efficient and rapid 3D shape regressor trained on generic RGB datasets. As an on-line tracking system, the identity of the unknown user is adapted on-the-fly resulting in improved 3D model reconstruction and consequently better tracking performance. The result is a robust RGBD face tracker, capable of handling a wide range of target scene depths, beyond those that can be afforded by traditional depth or RGB face trackers. Lastly, since the blendshape is not able to accurately recover the real facial shape, we use the tracked 3D face model as a prior in a novel filtering process to further refine the depth map for use in other tasks, such as 3D reconstruction.
\end{abstract}




\section{Introduction}

Tracking dynamic expressions of human faces is an important task, with recent methods~\cite{saragih11,cao14,weise11,bouaziz13,li13} achieving impressive results. However, difficult problems remain due to variations in camera pose, video quality, head movement and illumination, added to the challenge of tracking different people with many unique facial expressions.

\begin{figure}[!ht]
\centering
\subfigure[]{ \label{fig:intro1}\includegraphics[width=0.8in]{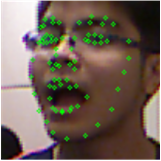} }
\subfigure[]{ \label{fig:intro4}\includegraphics[width=0.65in]{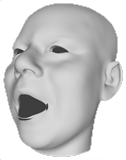} }
\subfigure[]{ \label{fig:intro2}\includegraphics[width=0.7in]{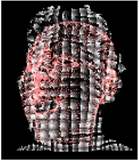} }
\subfigure[]{ \label{fig:intro3}\includegraphics[width=0.7in]{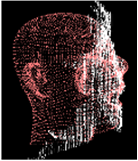} }
\caption{A tracking result of our proposed method. (a) The 3D landmarks projected to color frame. (b) The 3D blendshape. (c) The 3D frontal view, with the blendshape model in red and input point cloud in white. (d) The 3D side view.}
\label{fig:intro}
\end{figure}

Early work on articulated face tracking was based on Active Shape and Appearance Models~\cite{cootes95,cootes01,matt04} that fit a parametric facial template to the image. The facial template is learned from data and thus the tracking quality is limited by the amount of training samples and optimization method. Recently, alternative regression-based methods~\cite{BMVC07_booosted,cao12_cvpr,xiong13_cvpr,ren14,ECCV14_CoR} have resulted in better performance due to greater flexibility and computational efficiency.

Another common approach for face tracking is to use 3D deformable models as priors~\cite{cai10,ahlberg02,decarlo00,dornaika04,orozco2013,hai14}.
In general, a 3D face model is controlled by a set of shape deformation units. In the past, generic wireframe models (WFM)
were often employed for simplicity. However, WFM can only represent a coarse face shape and is insufficient for
fine-grained face tracking when dense 3D data is available.

Blendshape-based face models, such as the shape tensor used in the
FaceWareHouse~\cite{Cao:2013gy}, were developed for more
sophisticated, accurate 3D face tracking. By deforming dense 3D
blendshapes to fit facial appearances, facial motions can be estimated with high fidelity. Such techniques have gained attention recently due to the proliferation of consumer-grade range sensing devices, such as the Microsoft
Kinect~\cite{Kinect_2012}, which provide synchronized color (RGB)
images and depth (D) maps in real time. By integrating blendshapes
into dynamic expression models (DEM), several
approaches~\cite{weise11,bouaziz13,li13} have demonstrated
state-of-the-art tracking performance on RGBD input. It
can be observed that all of these tracking frameworks rely heavily
on the quality of input depth data. However, existing consumer-grade depth sensors tend to
provide increasingly unreliable depth
measurements when the objects are farther~\cite{Kinect_2012}. Therefore, these
methods~\cite{weise11,bouaziz13,li13} only work well at close
range, where the depth map retains fine structural details of the
face. In many applications, such as room-sized teleconferencing, the individuals tracked may be located at considerable
distances from the camera, leading to poor performance with existing methods.

One way of addressing depth sensor limitations is to use color as in~\cite{cao13,cao14}. These RGB-based
methods require extensive training to learn a 3D shape
regressor. The learned regressor serves as a prior for DEM
registration to 
RGB frames. Despite the high training cost, these
methods have tracking results comparable to RGBD-based
approaches.
Although RGB-only methods are not affected by inaccurate
depth measures, it is still challenging to track with high
fidelity at large object-camera distances. This is in
part due to reduced reliability of regression-based
updates at lower image resolutions, when there is less data for overcoming depth ambiguity.
Instead, we expect to achieve better tracking results if we were able to incorporate depth data while intelligently handling its inaccuracies at greater distances.


This motivates us to propose a robust RGBD face tracker 
combining the advantages of RGB regression and 3D point cloud
registration. Our contributions are as follows:
\begin{itemize}
\setlength{\itemsep}{0pt}
\item Our tracker is guided by a multi-stage 3D shape
regressor based on random forests and linear regression, which maps 2D image features back to blendshape parameters for a 3D face model.
This 3D shape regressor bypasses the problem of noisy depth data when obtaining a good initial estimate of the blendshape.


\item The subsequent joint 2D+3D optimization matches the facial blendshape
to both image and depth data robustly. This approach does not require an apriori blendshape model of the user, as shape parameters are updated on-the-fly.

\item Extensive experiments show that our 3D tracker performs robustly across a wide range of scenes and visual conditions, while maintaining or surpassing the tracking performance of other state-of-the-art trackers.

\item We use the DEM blendshape as a prior in a depth filtering process, further improving the depth map for fine 3D reconstruction.
\end{itemize}

The rest of this paper is organized as follows.
Section~\ref{sec:fw} outlines our proposed 3D face tracking
framework. Section~\ref{sec:reg} describes the 3D
shape regression in detail. DEM registration is further elaborated in
Section~\ref{sec:regis}. Section~\ref{sec:dr} describes our
depth recovery method using a 3D face prior.
Section~\ref{sec:exp} presents the experimental tracking and depth
recovery results.

\section{System Overview}
\label{sec:fw}
In this section we present the blendshape model that we use in this work, and our proposed tracking framework.
\subsection{The Face Representation}
We use the face models developed in the FaceWarehouse
database~\cite{Cao:2013gy}. As specified in~\cite{Cao:2013gy}, a
facial expression of a person can be approximated by
\begin{equation}
{V} = {C_r}{ \times _2} w_{id}^T { \times _3} w_{exp}^T
\label{eq_vCr}
\end{equation}
where $C_r$ is a 3D matrix (called reduced core tensor) of size ($N_v$,
$N_{id}$, $N_{e}$) (corresponding to number of vertices, number of
identities and number of expressions, respectively), $w_{id}$ is
an $N_{id}$-dimension identity vector, and $w_{exp}$ is an
$N_{e}$-dimension expression vector. \eqref{eq_vCr} basically
describes tensor contraction at the 2nd mode by $w_{id}$ and at
the 3rd mode by $w_{exp}$.

Similar to~\cite{cao13}, for real-time face tracking of one
person, given his identity vector ${w_{id}}$, it is more
convenient to reconstruct the $N_{e}$ expression blendshapes for
the person of identity ${w_{id}}$ as
\begin{equation}
{B_j} = {C_r}{ \times _2}w_{id}^T{ \times _3} u_{{\exp }_j}^T
\label{eq_bi}
\end{equation}
where $u_{{\exp }_j}$ is the pre-computed weight vector for the
$j$-th expression mode~\cite{Cao:2013gy}. In this way, an
arbitrary facial shape of the person can be represented as a
linear sum of his expression blendshapes:
\begin{equation}
V = {B_0} + \sum\limits_{j = 1}^{N_e-1} {({B_j} - {B_0}){e_j}}
\label{eq_sumbi}
\end{equation}
where $B_0$ is the neutral shape, and ${e_j} \in [0,1]$ is the
blending weight, $j = 1, \ldots, N_e-1$. Finally, a fully
transformed 3D facial shape can be represented as
\begin{equation}
S = R \cdot V(B,e) + T 
\label{eq_RTe}
\end{equation}
with the parameters $\theta = (R,T,e)$, where $R$ and $T$
respectively represent global rotation and translation,
and $e = \{ {e_j}\}$ defined in~\eqref{eq_sumbi} represent the
expression deformation parameters. In this work, we keep the 50 most significant identity knobs in the reduced core tensor $C_r$, hence ($N_v$, $N_{id}$, $N_{e}$) = (11510, 50, 47).

\subsection{Framework Pipeline}
\label{sec:fw:pipe}
\begin{figure*}[!ht]
\centering
\includegraphics[width=6.5in]{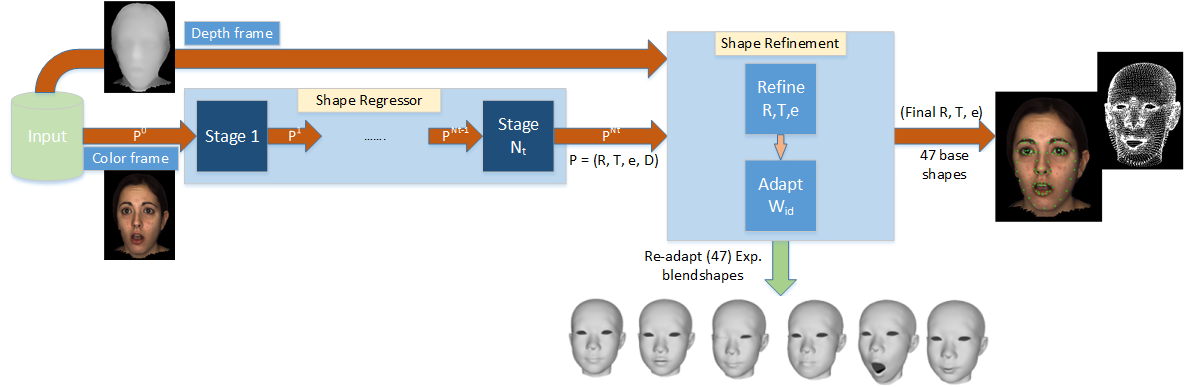}
\caption{The pipeline of the proposed method.}
\label{fig:pipeline}
\end{figure*}

Fig.~\ref{fig:pipeline} shows the pipeline of the proposed face tracking framework, which follows a coarse-to-fine multi-stage optimization design. In particular, our framework consists of two major stages: shape regression and shape refinement. The shape regressor performs the first optimization stage, which is learned from training data, to quickly estimate shape parameters $\theta$ from the RGB frame (cf. Section~\ref{sec:reg}). Then, in the second stage, a carefully designed optimization is performed on both the 2D image and the available 3D point cloud data to refine the shape parameters, and finally the identity parameter $w_{id}$ is updated to improve shape fitting to the input RGBD data (cf. Section~\ref{sec:regis}).

The 3D shape regressor is the key component to achieve our goal of 3D tracking at large distance, where quality of the depth map is often poor. Unlike the existing RGBD-based face tracking works, which either heavily rely on the accurate input point cloud (at close distances) to model shape transformation by ICP~\cite{weise11,bouaziz13} or use off-the-shelf 2D face tracker to guide the shape transformation~\cite{li13}, we predict the 3D
shape parameters directly from the RGB frame by the developed 3D regressor. This is motivated by the success of the 3D shape regression from RGB images used in~\cite{cao13,cao14}. The approach is especially meaningful for our considered large distance scenarios, where the depth quality is poor. Thus, we do not make use of the depth information in the 3D shape regression to avoid profusion of inaccuracies from the depth map.

Initially, a color frame $I$ is passed through the regressor to recover the shape parameters $\theta$. The projection of the $N_l$ ($N_l=73$) landmarks vertices of the 3D shape to image plane typically does not accurately match the 2D landmarks annotated in the training data. We therefore include 2D displacements $D$ in~\eqref{eq_RTe_err2d} into the parameter set and define a new global shape parameter set $P = ({\theta},D) = (R,T,e,D)$. The advantages of including $D$ in $P$ are two-fold. First, it helps train the regressor to reproduce the landmarks in the test image similar to those in the training set. Second, it prepares the regressor to work with unseen identity which does not appear in the training set~\cite{cao14}. In such case the displacement error $D$ may be large to compensate for the difference in identities. The regression process can be expressed as $ P^{out} = {f_{r}}(I,P^{in}) $, where $f_r$ is the regression function, $I$ is the current frame, $P^{in}$ and $P^{out}$ are the input (from the shape regression for the previous frame) and output shape
parameter sets, respectively. The coarse estimates $P^{out}$ are refined further in the next stage, using more precise energy optimization added with depth information. Specifically, $\theta = (R,T,e)$ are optimized w.r.t both the 2D prior constraints provided by the estimated 2D landmarks by the shape regressor and the 3D point cloud. Lastly, the identity vector $w_{id}$ is re-estimated given the current transformation.


\section{3D Shape Regression}
\label{sec:reg}

As mentioned in Section~\ref{sec:fw:pipe}, the shape regressor regresses over the parameter vector $P = (R,T,e,D)$. To train the regressor, we must first recover these parameters from training samples, and form training data pairs to provide to the training algorithm. In this work, we use the face databases from~\cite{LFWTech,gtav,Cao:2013gy} for training.

\subsection{Shape Parameters Estimation from Training Data}
We follow the parameter estimation process in~\cite{cao13}.
Denoting $\Pi _p$ the camera projection function from 3D world
coordinates to 2D image coordinates, $(R, T, w_{id}, w_{exp})$ are
first extracted by minimizing the 2D errors in each sample:
\begin{equation}
\mathop {\min }\limits_{R,T,{w_{id}},{w_{\exp }}} {\sum\limits_{i
= 1}^{N_l} {\left\| {{\Pi _p}\left( {R{{\left( {{C_r}{ \times
_2}w_{id}^T{ \times _3}w_{\exp }^T} \right)}_i} + T} \right) -
{l_i}} \right\|} ^2} \label{eq_err2d1}
\end{equation}
where $\left\{ {{l_i}|i = 1, \ldots, N_l } \right\}$ are the
ground truth landmarks of the training data and $N_l=73$. Note
that $w_{exp}$ \textit{will be discarded} since we only need
$w_{id}$ to generate the individual expression blendshapes
$\left\{ {{B_j}} \right\}$ of the current subject as
in~\eqref{eq_bi} for later optimization over (R,T,e).

With the initially extracted parameters in~\eqref{eq_err2d1}, we
refine $w_{id}$ by alternatingly optimizing over $w_{id}$ and
(R,T,$w_{exp}$). Particularly, we first keep (R, T, $w_{exp}$)
fixed for each sample, and optimize over $w_{id}$ across all the
samples of the same subject:
\begin{equation}
\mathop {\min }\limits_{{w_{id}}} \sum\limits_{k = 1}^{N_s}
{\sum\limits_{i = 1}^{N_l} {{{\left\| {{\Pi _p}\left(
{{R_k}{{\left( {{C_r}{ \times _2}w_{id}^T{ \times _3}w_{{k_{\exp
}}}^T} \right)}_i} + {T_k}} \right) - {l_{{k_i}}}} \right\|}^2}} }
\label{eq_wid}
\end{equation}
where $N_s$ denotes the total number of training samples for the
same subject. Then for each sample we keep $w_{id}$ fixed and
optimize over (R, T, $w_{exp}$) as in~\eqref{eq_err2d1}. This
process is repeated until convergence. We empirically observe that
running the above process for three iterations gives reasonably
good results. We then can generate user-specific blendshapes
$\left\{ {{B_i}} \right\}$ as in~\eqref{eq_bi}.

Finally, we recover the expression weights $e$ by minimizing the
2D error over (R,T,e) again:
\begin{equation}
\mathop {\min }\limits_{R,T,e} \sum\limits_{i = 1}^{N_l}
{{{\left\| {{D_i}} \right\|}^2}} \label{eq_RTe_err2d}
\end{equation}
where ${D_i} = {\Pi _p}\left( {{S_i}} \right) - {l_i}$ and $S_i$
is a 3D landmark vertex of the blendshape corresponding to $l_i$.
From~\eqref{eq_RTe_err2d}, we also obtain the 2D displacement
vector $D = \left\{ {{D_i}} \right\}$ as a by-product. Eventually,
following~\cite{cao14}, for each training data sample, we generate
a number of guess-truth pairs $\left\{ {{I_i},P_i^0,P_i^g} \right\}$,
where the guessed vector $P_i^0$ is produced by randomly
perturbing the ground truth parameters $P_i^g$ extracted through
the above optimization. In this way, we create $N$ training pairs
in total.

\subsection{Shape Regression Training}
Given the training pairs from the previous section, we follow the
feature extraction and shape regression method in~\cite{ren14},
which combines local binary features extracted using the trained
random forests of all the landmarks. The local binary features are
aggregated into a global feature vector which is then used to
train a linear regression model to predict the shape parameters.
In our work, we train the regressor to predict $(R,T,e,D)$
simultaneously, directly from the input RGB frame in contrast to~\cite{ren14} where the regressor simply updates only the 2D displacements.

Algorithm~\ref{algo:train} shows the detailed training procedure.
In particular, we calculate the 2D landmark positions from the
shape parameters, and for each landmark $l_i$, we randomly sample
pixel intensity-difference features~\cite{cao12_cvpr} within a
radius $r_i$. These pixel-difference features are then used to
train a random forest ${Forest}_i$. For every training sample
$M_k$, we pass it through the forest and recover a binary vector
$F_{k,i}$ which has the length equal to the number of leaf nodes
of the forest. Each node that responds to the sample will be
represented as 1 in $F_{k,i}$; otherwise it will be 0. The local
binary vectors from $N_l$ landmarks are concatenated to form a global
binary vector ${\Phi _k}$ representing the training sample $k$.
Then, the global binary feature vectors are used to learn a global
linear regression matrix $\boldsymbol{W}$ which predicts the
updating shape parameters $\Delta P$ from those binary global
vectors. After that, the guessed shape parameters are updated and
enter the next iteration.

\begin{algorithm}
\caption{The regressor training algorithm} \label{algo:train}
\KwData{N training samples $M_{k} = \left\{ {{I_k},P_k^0,P_k^g}
\right\}$} \KwResult{The shape regressor} \For{$t \leftarrow 1$
\KwTo $N_t$} {
    \For{$i \leftarrow 1$ \KwTo $N_l$} {
        $Forest_i$ $\longleftarrow$ TrainForest($l_i$)\;
        \For{$k \leftarrow 1$ \KwTo $N$} {
            $F_{k,i}$ $\longleftarrow$ Pass($M_k$, $Forest_i$)\;
        }
    }
    \For{$k \leftarrow 1$ \KwTo $N$} {
        ${\Phi ^t}\left( {{I_k},P_k^{t - 1}} \right) \leftarrow concat\left( {{F_{k,i}}} \right)$\;
    }
$ \min \sum\limits_{k = 1}^N {{{\left\| {\Delta P_k^t - {W^t}{\Phi
^t}\left( {{I_k},P_k^{t - 1}} \right)} \right\|}^2} + \lambda
{{\left\| {{W^t}} \right\|}^2}} $\;
    \For{$k \leftarrow 1$ \KwTo $N$} {
        $P_k^t \leftarrow P_k^{t - 1} + {W^t}{\Phi ^t}\left( {{I_k},P_k^{t - 1}} \right)$\;
    }
}
\end{algorithm}

Similar to~\cite{ren14}, we let the regressor learn the best search radius $r_i$ during training. The training face samples have been normalized to the size of approximately 120x120 pixels, about the same size as the face captured by Kinect at 0.7m distance. Thus at runtime, we simply rescale the radius inversely proportional to the current z-translation $T_z$.

\section{3D Shape Refinement}
\label{sec:regis}

At this stage, we refine the shape parameters using both RGB and
depth images, and also update the shape identity. Specifically, $(R,T)$ and $e$ are alternatingly optimized. After convergence, the identity vector $w_{id}$ is updated based on the final shape parameters vector $\theta = (R,T,e)$.




\subsection{Facial Shape Expressions and Global Transformation}
\label{sec:refineRTe}
We simultaneously refine $(R,T,e)$ by
optimizing the following energy:
\begin{equation}
R,T,e = \arg \min {E_{2D}} + {\omega _{3D}}{E_{3D}} + {E_{reg}}
\label{eq_refineRTe}
\end{equation}
where $\omega _{3D}$ is a tradeoff parameter, $E_{2D}$ is the 2D
error term measuring the 2D displacement errors, $E_{3D}$ is the
3D ICP energy term measuring the geometry matching between the 3D
face shape model and the input point cloud, and $E_{reg}$ is the
regularization term to ensure the shape parameter refinement is
smooth across the time. Particularly, $E_{2D}$, $E_{3D}$ and
$E_{reg}$ are defined as
\begin{eqnarray}
{E_{2D}} = \frac{1}{{N_l}}{\sum\limits_{i = 1}^{N_l} {\left\|
{{\Pi _p}\left( {{S_i}\left( {R,T,e} \right)} \right) - {l_i}}
\right\|} ^2} \label{eq_e2d} \\
{E_{3D}} = \frac{1}{{{N_d}}}\sum\limits_{k = 1}^{{N_d}} {{{\left(
{\left( {{S_k}\left( {R,T,e} \right) - {d_k}} \right) \cdot {n_k}}
\right)}^2}} \label{eq_e3d} \\
{E_{reg}} = \alpha {\left\| {\theta - {\theta ^*}} \right\|^2} +
\beta {\left\| {\theta  - 2{\theta ^{(t - 1)}} + {\theta ^{(t -
2)}}} \right\|}^2 . \label{eq_ereg}
\end{eqnarray}

In~\eqref{eq_e2d}, the tracked 2D landmarks \{$l_i$\} are computed
from the raw shape parameters $(R,T,e,D)$, which are usually
quite reliable. In~\eqref{eq_e3d}, $N_d$ is the number of ICP
corresponding pairs that we sample from the blendshape and the
point cloud, and $d_k$ and $n_k$ denote point $k$ in the point
cloud and its normal, respectively. By minimizing $E_{3D}$,
we essentially minimize the point-to-plane ICP distance between
the blendshape and the point cloud~\cite{klow04}. This is to help
slide the blendshape over the point cloud to avoid local minima
and recover a more accurate pose. In~\eqref{eq_ereg}, $\theta ^*$ is
the raw output $(R,T,e)$ from the shape regressor,
$\theta ^{(t-1)}$ and $\theta ^{(t-2)}$ are the shape parameters
from the previous two frames, and $\alpha$ and $\beta$ are
tradeoff parameters. The two terms in~\eqref{eq_ereg} 
represent a data fidelity term and a Laplacian smoothness term.

In our implementation, we iteratively optimize over the global
transformation parameters $(R,T)$ and the local deformation
parameter $e$, which leads to faster convergence and lower
computational cost. In the $(R,T)$ optimization, $\omega _{3D}$ is
set to 2; $\alpha$, $\beta$ are set to 100 and 10000 for $R$, 0.1
and 10 for $T$, respectively. For optimization over $e$, $\omega
_{3D}$ is set to 0.5; $\alpha$ and $\beta$ are both set to zero so
as to maximize spontaneous local deformations. The non-linear
energy function is minimized using the ALGLIB::BLEIC bounded
solver\footnote{http://www.alglib.net/} to keep $e$ in the valid range of [0,1].

\begin{figure}[!ht]
\centering
\subfigure[]{\label{fig:refine1} \includegraphics[width=0.7in]{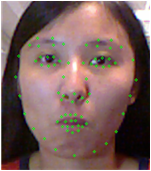} }
\subfigure[]{\label{fig:refine2} \includegraphics[width=0.7in]{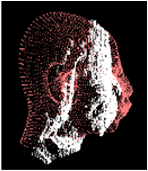} }
\subfigure[]{\label{fig:refine3} \includegraphics[width=0.7in]{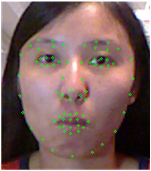} }
\subfigure[]{\label{fig:refine4} \includegraphics[width=0.7in]{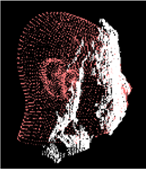} }
\caption{The effect of $E_{3D}$ term. (a,b): The result without using $E_{3D}$. (c,d): The result using $E_{3D}$. Notice the displacement between
the point cloud and the model, as well as the over-deformation of
the mouth in (b).}
\label{fig:refine}
\end{figure}

Fig.~\ref{fig:refine} gives an example to show the effect of the
$E_{3D}$ term. We can see that for the result without using
$E_{3D}$, there is a large displacement between the point cloud
and the model and there is also noticeable over-deformation of the
mouth. This demonstrates that without using the 3D information,
the 2D tracking may appear fine yet the actual 3D transformation is largely
incorrect.

\subsection{Updating Shape Identity}
In the last step, we refine the identity vector to better adapt
the expression blendshapes to the input data. We solve for
$w_{id}$ by minimizing the following objective function:

\begin{equation}
{w_{id}} = \arg \min {E'_{2D}} + {\omega _{3D}}{E'_{3D}}
\end{equation}
where
\begin{equation}
\begin{aligned}
%
&{E'_{2D}} = \frac{1}{{N_l}}\sum\limits_{i = 1}^{N_l} {{{\left\|
{{\Pi _p}\left( {R{{\left( {{C_r}{ \times _2}w_{id}^T{ \times
_3} {\gamma ^T}} \right)}_i} + T} \right) - {l_i}}
\right\|}^2}} \\
&E'_{3D} = \frac{1}{{{N_d}}}\sum\limits_{k = 1}^{{N_d}} {{{\left\|
{R{{\left( {{C_r}{ \times _2}w_{id}^T{ \times _3}{\gamma ^T}}
\right)}_k} + T - {d_k}} \right\|}^2}} \label{eq_e_wid}
\end{aligned}
\end{equation}
with ${\gamma} = (1- \sum\limits_{j = 1}^{N_e-1} e_j) u_{{\exp
}_0} + \sum\limits_{j = 1}^{N_e-1} e_j u_{{\exp }_j}$.

Note that $E'_{3D}$ is the point-to-point ICP energy and it behaves slightly differently from $E_{3D}$ in~\eqref{eq_e3d}. Minimizing $E'_{3D}$ helps align the blendshape to the point cloud in a more direct way on the surface to recover detailed facial characteristics.

In our experiments, we empirically set $\omega _{3D}$ to 0.5,
meaning that we give more weight to the 2D term to encourage the
face model to fit closer to the tracked landmarks, especially the face countour. Gradient-based optimizations such as BFGS are ineffective toward this energy, and thus we run one iteration of coordinate descent at each frame to stay within the computational budget. We find that
$w_{id}$ usually converges in under 10 frames after tracking starts.
To save computational time, we set a simple rule in which updating
identity stops either after $w_{id}$ converges or after 10 frames.

\begin{figure}
\centering \subfigure[]{\label{fig:adapt1}
\includegraphics[width=0.5in]{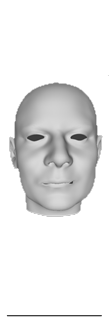} }
\subfigure[]{\label{fig:adapt2}
\includegraphics[width=0.52in]{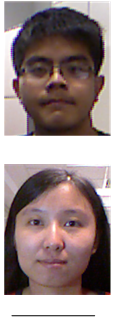} }
\subfigure[]{\label{fig:adapt3}
\includegraphics[width=1.8in]{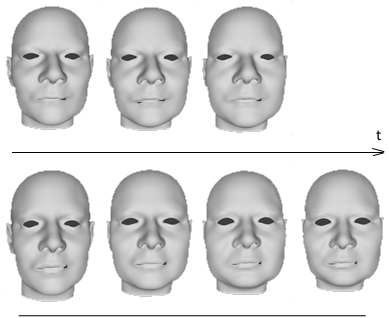} } 
\caption{Adapting identity over time. (a) The common initial base shape. (b)
Appearances of two testers. (c) For the male tester, the identity
parameter $w_{id}$ converges after three frames, compared to the female tester's four frame convergence.} \label{fig:adapt}
\end{figure}

Fig.~\ref{fig:adapt} shows some results on adapting the identity
parameter over time. After a few
iterations of updating $w_{id}$, the face model fits significantly better to each individual subject.

\section{Depth Recovery with Dense Face Priors}
\label{sec:dr}
In this section, we further develop one application to show the
usefulness of the final blendshape model for each frame, i.e.
using the dense blendshape model as the prior for depth recovery.
Although the final blendshape itself is a good approximation to
the real face and sufficiently good for the tracking purpose, it
might not be sufficient for other applications such as the 3D face
reconstruction. Thus, it is meaningful to use face priors to
refine noisy depth maps. Existing methods for depth
recovery~\cite{WMF_TIP12,RGBZ_EUG12,AR_ECCV12,Zhao_ICME13,Chen2015} usually utilize general prior information such as piece-wise smoothness and the
corresponding color guidance, and thus they tend to produce a
plane-like surface. To address these deficiencies, the use of semantic priors has also been considered, e.g., rigid object
priors~\cite{CVPR13_PriorRec} and non-rigid face
priors~\cite{ACCV14_FaceDepth}, for 3D reconstruction and depth
recovery.

Our work is based on~\cite{ACCV14_FaceDepth} but extends it in several significant ways. \cite{ACCV14_FaceDepth} mainly introduces
the idea of using the face prior and focuses on the depth recovery of
one single RGBD image with the help of face registration. It uses
a coarse generic wireframe face model, which can only provide a
limited reliable depth prior. In contrast, we employ
our optimized final blendshape model which can provide dense prior
information. We also incorporate depth recovery with real-time face
tracking, for which we develop a local filtering based depth
recovery for fast processing.

In particular, similar to~\cite{ACCV14_FaceDepth}, the recovery of
depth map $X$ is formulated as the following energy minimization
problem:
\begin{equation}
\min\limits_{X} E_r(X) + \lambda_d E_d(X,Z) + \lambda_f E_f(X, V),
\label{equ:DepthRec}
\end{equation}
where the smoothness term $E_r(X)$ measures the quadratic
variations between neighboring depth pixels, the fidelity term
$E_d(X,Z)$ is adopted to ensure $X$ does not significantly depart from
the depth measurement $Z$, and the face prior term $E_f(X,V)$
utilizes the blendshape prior $V$ to guide the depth recovery. We
define
\begin{equation}
E_r(X) =  \frac{1}{2} \sum\limits_{i} \sum\limits_{j\in\Omega_i}
\alpha_{ij} ( X(i) - X(j) )^2 , \label{equ:Er}
\end{equation}
where $i$ and $j$ represent the pixel index, $\Omega_i$ is
the set of neighboring pixels of pixel $i$, $\alpha_{ij}$ is the
normalized joint trilateral filtering (JTF) weight which is
inversely proportional to pixel distance, color difference, and
the depth difference~\cite{ACCV14_FaceDepth}. For the fidelity
term $E_d(X,Z)$, we use the Euclidean distance between $X$ and $Z$,
i.e., $E_d(X,Z) = \frac{1}{2} \|X-Z\|^2$. For simplicity, we use
$V$ to represent the depth map generated by rendering the current
3D blendshape model at the color camera viewpoint. Then, the face
prior term $E_f(X,V)$ is computed as $E_f(X,V) = \frac{1}{2}
\|X-V\|^2$.

A simple recursive solution to (14) is obtained by the vanishing gradient condition, resulting in
\begin{equation}
    \displaystyle X^{t+1}(i) =
\frac{\lambda_d Z(i) + \lambda_f V(i) + \displaystyle
\sum_{j\in\Omega_i}(\alpha_{ij}+\alpha_{ji})X^t(j)}
 {\lambda_d+\lambda_f + \displaystyle\sum_{j\in\Omega_i}(\alpha_{ij}+\alpha_{ji})},
\label{equ:Jacobi_Iter}
\end{equation}
where the superscript represents the number of iterations. Such
filtering process is GPU-friendly and the number of iterations can
be explicitly controlled to achieve a better trade-off between
recovery accuracy and speed.

\section{Experiments}
\label{sec:exp}
\subsection{Tracking Experiments}
\label{sec:exp:track}
We carried out extensive tracking experiments on synthetic BU4DFE sequences and real videos captured by a Kinect camera. We compared the tracking performance of our method to that of RGB-based trackers DDER\cite{cao14}, CoR\cite{ECCV14_CoR} and RLMS\cite{saragih11} in terms of average root mean square error (RMSE) in pixel positions of 2D landmarks. In the tracking context, we evaluated trackers' robustness by comparing the proportions of unsuccessfully tracked frames.

\subsubsection{Evaluations on Synthetic Data}
\label{sec:exp:track:bu4d}
The BU4DFE dataset~\cite{BU4DFE_08} contains sequences of high-resolution 3D dynamic facial expressions of human subjects. We rendered these sequences into RGBD to simulate the Kinect camera~\cite{Kinect_2012} at three distances: 1.5m, 1.75m and 2m with added rotation and translation. In total, we collected tracking results from 270 sequences. The dataset does not provide ground truth, so we used the RLMS tracker~\cite{saragih11}, which works well on BU4DFE sequences, to recover 2D landmarks on the images rendered at 0.6m, which were then reprojected to different distances and treated as ground truth.

The overall evaluation results are shown in Table~\ref{tbl:bu4d_eval}. Our tracker performed comparably to the state-of-the-art CoR~\cite{ECCV14_CoR} and outperformed the blendshape-based DDER~\cite{cao14}. CoR did not produce results for sequences at 1.75m and 2m, with the faces too small for it to handle.

\begin{table}[ht]
\centering
\caption{Evaluation results of the proposed method and other face trackers on BU4D dataset. RMSE is measured in pixels.}
\begin{tabular}{|c|c|c|c|c|}
\hline 
Dataset   		& DDER~\cite{cao14} &  CoR~\cite{ECCV14_CoR}  &  Ours 	\\\hline
BU4D (1.5 m) 	& 	2.20	& 	\textbf{1.05}	& 1.27	\\\hline
BU4D (1.75 m)   &  	1.94	&  	n/a		& \textbf{1.14}	\\\hline
BU4D (2.0 m)	&  	1.76	&  	n/a		& \textbf{1.14}	\\\hline
\end{tabular}
\label{tbl:bu4d_eval}
\end{table} 

\begin{figure}[!ht]
\centering
\includegraphics[width=3.3in]{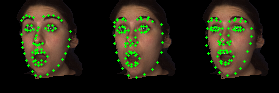}
\caption{A sample from BU4DFE dataset, rendered at 1.5m. From left to right: results by CoR, DDER and our tracker.}
\label{fig:bu4d_compare}
\end{figure}

\subsubsection{Experiments on Real Data}
\label{sec:exp:track:real}
We compared the tracking performance of our approach to other methods on 11 real sequences at various distances, with different lighting conditions, complex head movements as well as facial expressions. We used RLMS to recover the ground truth, and manually labeled the frames that were incorrectly tracked.

\begin{figure*}[!ht]
\centering
\includegraphics[width=7.0in]{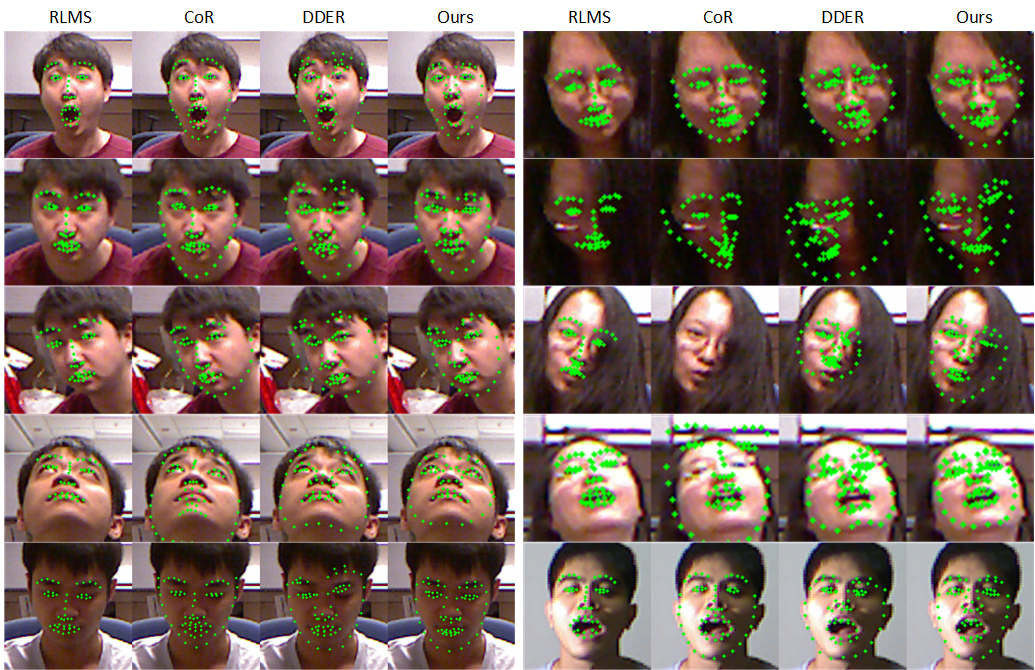}
\caption{Each group of four shows results of four trackers on the same frame. From left to right: RLMS, CoR, DDER and our method. Our tracker and RLMS can handle occlusion by hair. In general, our tracker is robust to large rotation and it models realistic facial deformations.}
\label{fig:real_compare}
\end{figure*}

\begin{figure*}[!ht]
\centering
\subfigure[]{\label{fig:occ1}\includegraphics[width=1.1in]{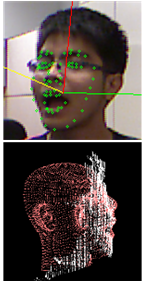}}
\subfigure[]{\label{fig:occ2}\includegraphics[width=1.1in]{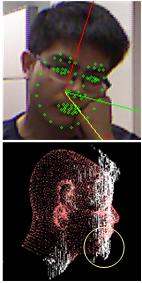}}
\subfigure[]{\label{fig:occ3}\includegraphics[width=1.1in]{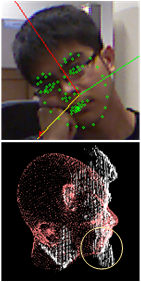}}
\subfigure[]{\label{fig:occ4}\includegraphics[width=1.1in]{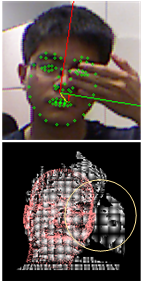}}
\subfigure[]{\label{fig:occ5}\includegraphics[width=1.1in]{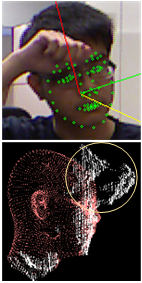}}
\subfigure[]{\label{fig:occ6}\includegraphics[width=1.1in]{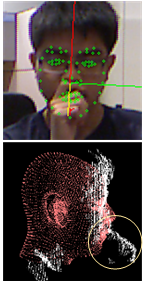}}
\caption{Example showing that the proposed tracker can handle partial occlusion of the face. The first row shows the resulting projected landmarks and the head orientation as 3 axes (red, green, yellow axes are yaw, pitch and roll, respectively). The second row shows the 3D view of the blendshape model (in red) and the input point cloud (in white) of each corresponding frame. Except (c) where the frontal view is shown, (a,b,c,e,f) show the side view. In each frame, the occlusion on the point cloud is circled in yellow. Tracking performance is not measured for this video and it is not included in Table~\ref{tbl:real_eval}, because we recorded this sequence after we had finished all the benchmarks.}
\label{fig:occlusion}
\end{figure*}

The results are shown in Table~\ref{tbl:real_eval}. For RLMS, we only considered the performance on frames that had been manually labeled, since its results were otherwise used as ground truth. Note that the inclusion of RLMS is mainly used as a reference and does not reflect its true performance, as only incorrectly tracked frames were measured. Once again, our method outperformed DDER and was very close to CoR. The consistent error values demonstrated that our tracker is stable, particularly under large rotations or when the face is partially covered, as illustrated in Fig.~\ref{fig:real_compare} and Fig.~\ref{fig:occlusion}.

To better assess the robustness of each tracker, we compared the percentage of aggregated lost frames from all sequences in Table~\ref{tbl:track_perc}. The mistracked frames were decided either by empty output, or by large RMSE ($RMSE > \tau$, with $\tau=10$). We also did not count sequences \textit{luc03} for DDER, nor \textit{luc03} and \textit{luc04} for CoR, toward their overall percentages because the faces were not registered correctly from the beginning, which was perhaps largely due to the face detector failing to locate the face correctly. This showed that the 2D+3D optimization combination of our method provides robust tracking overall.

\begin{table}[ht]
\centering
\caption{Evaluation results of the proposed method and other face trackers on real videos. RMSE is measured in pixels.}
\begin{tabular}{|c|c|c|c|c|}
\hline 
Dataset 	& DDER~\cite{cao14} &  CoR~\cite{ECCV14_CoR}  & RLMS~\cite{saragih11}  & Ours 	\\\hline
dt01  		& 9.65	& 	4.15	&	6.04	& 4.51	\\\hline
ar00  		& 3.41	& 	66.72	&	7.41	& 2.36	\\\hline
dt00  		& 3.57	& 	1.65	&	4.63	& 2.29	\\\hline
my01  		& 5.61	& 	2.79	&	4.35	& 2.89	\\\hline
fw01  		&  6.5	& 	3.27	&	36.11	& 4.85	\\\hline
fw02  		& 5.34	& 	1.80	&	2.56	& 3.50	\\\hline
luc01  		& 4.96	& 	2.38	&	5.86	& 3.49	\\\hline
luc02  		& 3.95	& 	1.51	&	2.04	& 3.02	\\\hline
luc03 (2m)  & 37.17	& 	n/a		&	1.67	& 1.77	\\\hline
luc04 (2m) 	& 2.63	& 	62.45	&	n/a		& 1.84	\\\hline
luc05  		& 3.39	& 	2.39	&	3.44	& 2.88	\\\hline
\end{tabular}
\label{tbl:real_eval}
\end{table}    

\begin{table}[ht]
\centering
\caption{The overall percentage of lost frames during tracking from all real videos.}
\begin{tabular}{|c|c|c|c|}
\hline
DDER~\cite{cao14} &  CoR~\cite{ECCV14_CoR}  & RLMS~\cite{saragih11}  & Ours 	\\\hline
2.21\%	&	7.22\%	&	3.61\%	&	\textbf{0.74\%}	\\\hline
\end{tabular}
\label{tbl:track_perc}
\end{table}

\subsubsection{Running Time}
\label{sec:exp:time}
Our tracker is implemented in native C++, parallelized with TBB\footnote{https://www.threadingbuildingblocks.org/}, with the GPU only used for calculating the 47 base expression blendshapes in~\eqref{eq_bi}. Running time was measured on a 3.4GHz Core-i7 CPU machine with a Geforce GT640 GPU. Shape regression ran in 5ms, refining (R,T,e) took 12ms, with auxiliary processing taking another 10ms. Overall, without identity adaptation, the tracker ran at 30Hz. The bottleneck is in optimizing for $w_{id}$ which took 14ms, while calculating 47 base blendshapes took 80ms on the GT640 GPU with 384 CUDA cores. This process is only carried out at initialization or during tracker restarts. The use of modern GPU cards with higher CUDA core counts should remove this bottleneck.

\subsection{Depth Recovery Experiments}
\label{sec:exp:depth}
\subsubsection{Synthetic Data}
\label{sec:exp:depth:bu4d}
We used the same set of BU4DFE sequences as in section~\ref{sec:exp:track:bu4d} at 1.75m and 2m. Instead of evaluating the tracking accuracy, we measured the surface reconstruction error with respect to the 3D synthetic surface used for generating the data. To simulate different depth ranges of the target, we increased the noise level of the input depth map according to~\cite{Kinect_2012}. We ran the tracker on these sequences and collected the surface of the blendshape (BSSurface) as well as the enhanced depth map, which was filtered using face priors (DRwFP). We compared these two surfaces to the ground truth surface. Additionally, we compared our method to the depth recovery method in~\cite{Chen2015} using Mean Absolute Error (MAE) $\frac{1}{{\left| \Omega  \right|}}\sum\limits_{i \in \Omega } {\left| {{d_i} - {g_i}} \right|}$ in mm, where $\Omega$ is the set of valid depth pixels, while $d_i$ and $g_i$ are the recovered and ground truth depth values respectively. The results are summarized in Table~\ref{tbl:depth_bu4d}.

The results show that the high noise levels, often higher than that of the actual Kinect depth data, led to large errors in blendshape modeling. However, the face guidance filter mitigated these problems and recovered depth maps that were closer to the ground truth surface. The improvements range from $15\%$ at 1.75m to $20\%$ at 2m better than~\cite{Chen2015}.

\begin{table}[ht]
\centering
\caption{Average MAE in mm of depth reconstruction on BU4DFE dataset.}
\begin{tabular}{|c|c|c|c|}
\hline 
Dataset   		& ~\cite{Chen2015} &  BSSurface  &  \textbf{DRwFP} 	\\\hline
BU4DFE (1.75 m) &   2.83	&  	9.16	& \textbf{2.39} 	\\\hline
BU4DFE (2.0 m)	&  	3.59	&  	8.79	& \textbf{2.85} 	\\\hline
\end{tabular}
\label{tbl:depth_bu4d}
\end{table} 

\subsubsection{Real Data}
\label{sec:exp:depth:real}
\begin{figure}[!ht]
\centering
\subfigure[]{\label{fig:depthrec1}\includegraphics[width=0.5in]{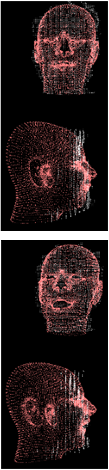}}
\subfigure[]{\label{fig:depthrec2}\includegraphics[width=0.89in]{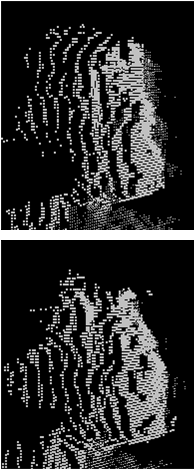}}
\subfigure[]{\label{fig:depthrec3}\includegraphics[width=0.89in]{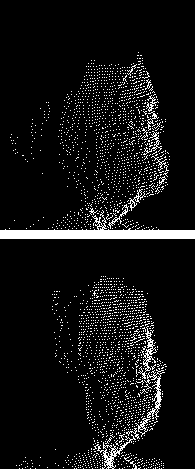}}
\subfigure[]{\label{fig:depthrec4}\includegraphics[width=0.89in]{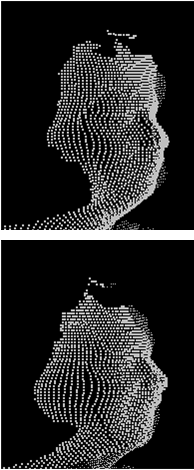}}
\caption{Depth recovery on real depth maps. 
(a) The blendshape priors
(b) The raw depth maps
(c) The depth maps refined by~\cite{Chen2015}
(d) The depth maps refined with face priors.
}
\label{fig:depthrec}
\end{figure}

As we do not have ground truth for real data, in this section we only provide visual results of the recovered depth map at 2m.
Fig.~\ref{fig:depthrec} shows depth recovery results on two sample depth frames. It is difficult to recognize any facial characteristics from the raw depth maps. The filter in~\cite{Chen2015} smoothed out the depth maps but was not able to recover any facial details. In contrast, our depth filter with face priors was able to reconstruct the facial shapes with recognizable quality.

\section{Conclusion}
\label{sec:conclude}
We presented a novel approach to RGBD face tracking, using 3D facial blendshapes to simultaneously model the head movements as well as facial expressions. The tracker is driven by a fast shape regressor, which allows the tracker to perform consistently at any distance, beyond the working range of current state-of-the-art RGBD face trackers. This 3D shape regressor directly estimates shape parameters together with 2D landmarks in the input color frame. The shape parameters are refined further by optimizing a well-designed 2D+3D energy function. Using this framework, our tracking can automatically adapt the 3D blendshapes to better fit the individual facial characteristics of tracked humans. Through extensive experiments on synthetic and real RGBD videos, our tracker performed consistently well in complex conditions and at different distances.

With the ability to model articulated facial expressions and complex head movements, our tracker can be deployed in various tasks such as animation and virtual reality. In addition, we use the blendshape as a prior in a novel depth filter to better reconstruct the depth map, even at larger distances. The refined depth map can later be used together with the blendshape to reproduce the facial shape regardless of object-camera distances.

\section*{Acknowledgements}
We specially thank Deng Teng for his kind help in collecting the real test sequences used in this paper.

{
\bibliographystyle{ieee}
\bibliography{egbib}
}

\end{document}